\documentclass{article}

\PassOptionsToPackage{numbers, compress}{natbib}
\usepackage[preprint]{neurips_data_2024}
\usepackage[numbers]{natbib}
\usepackage[utf8]{inputenc} % allow utf-8 input
\usepackage[T1]{fontenc}    % use 8-bit T1 fonts
\usepackage{hyperref}       % hyperlinks
\usepackage{url}            % simple URL typesetting
\usepackage{booktabs}       % professional-quality tables
\usepackage{amsfonts}       % blackboard math symbols
\usepackage{nicefrac}       % compact symbols for 1/2, etc.
\usepackage{microtype}      % microtypography
\usepackage{xcolor}         % colors
\usepackage{graphicx}
\usepackage{multirow}
\usepackage{colortbl}
\usepackage{caption}
\usepackage{CJKutf8}

\newcommand{\ie}{\textit{i}.\textit{e}., }

\newcommand{\etc}{\textit{etc}. }
\newcommand{\cc}[1]{\begin{CJK*}{UTF8}{gbsn}#1\end{CJK*}}

\definecolor{cadmiumgreen}{rgb}{0.0, 0.42, 0.24}
\definecolor{cadmiumred}{rgb}{0.89, 0.0, 0.13}

\definecolor{darkergreen}{RGB}{21, 152, 56}

\title{CFinBench: A Comprehensive Chinese Financial Benchmark for Large Language Models}

\author{
	Ying Nie\textsuperscript{\rm 1}$^*$, Binwei Yan\textsuperscript{\rm 1}\thanks{Equal Contribution.}, Tianyu Guo\textsuperscript{\rm 1}, Hao Liu\textsuperscript{\rm 1}, Haoyu Wang\textsuperscript{\rm 1}, Wei He\textsuperscript{\rm 1}, \\
	\textbf{Binfan Zheng\textsuperscript{\rm 2}, Weihao Wang\textsuperscript{\rm 3}, Qiang Li\textsuperscript{\rm 3}, Weijian Sun\textsuperscript{\rm 2}, Yunhe Wang\textsuperscript{\rm 1}\thanks{Corresponding Author.}, Dacheng Tao\textsuperscript{\rm 4}} \\
	\textsuperscript{\rm 1}Huawei Noah’s Ark Lab ~~~\textsuperscript{\rm 2}Huawei GTS ~~~\textsuperscript{\rm 3}Huawei Group Finance \\
	\textsuperscript{\rm 4}Nanyang Technological University \\
	\texttt{\small \{ying.nie, yanbinwei, yunhe.wang\}@huawei.com} ~~\texttt{\small \{dacheng.tao\}@gmail.com}\\
}
\begin{document}

\maketitle

\begin{abstract}
Large language models (LLMs) have achieved remarkable performance on various NLP tasks, yet their potential in more challenging and domain-specific task, such as finance, has not been fully explored. In this paper, we present CFinBench: a meticulously crafted, the most comprehensive evaluation benchmark to date, for assessing the financial knowledge of LLMs under Chinese context. In practice, to better align with the career trajectory of Chinese financial practitioners, we build a systematic evaluation from 4 first-level categories: (1) \emph{Financial Subject}: whether LLMs can memorize the necessary basic knowledge of financial subjects, such as economics, statistics and auditing. (2) \emph{Financial Qualification}: whether LLMs can obtain the needed financial qualified certifications, such as certified public accountant, securities qualification and banking qualification. (3) \emph{Financial Practice}: whether LLMs can fulfill the practical financial jobs, such as tax consultant, junior accountant and securities analyst. (4) \emph{Financial Law}: whether LLMs can meet the requirement of financial laws and regulations, such as tax law, insurance law and economic law. CFinBench comprises 99,100 questions spanning 43 second-level categories with 3 question types: single-choice, multiple-choice and judgment. We conduct extensive experiments of 50 representative LLMs with various model size on CFinBench. The results show that GPT4 and some Chinese-oriented models lead the benchmark, with the highest average accuracy being 60.16\%, highlighting the challenge presented by CFinBench. The dataset and evaluation code are available at \url{https://cfinbench.github.io/}.

\end{abstract}

\section{Introduction}
Recently, there has been a significant advancement in LLMs, exemplified by the representative models like ChatGPT~\cite{chatgpt}, GPT4~\cite{openai2023gpt4}, LLaMA~\cite{touvron2023llama,touvron2023llama2,llama3}, Baichuan~\cite{yang2023baichuan}, InternLM~\cite{team2023internlm} and ChatGLM~\cite{zeng2022glm}, \etc At the same time, the corresponding evaluation works for LLMs are flourishing and a series of evaluation benchmarks have been proposed like MMLU~\cite{hendrycks2020measuring}, C-Eval~\cite{huang2024c}, Xiezhi~\cite{gu2024xiezhi} and AGIEval~\cite{zhong2023agieval}, \etc These benchmarks have been instrumental in catalyzing the progress of LLMs, as they enable the quantitative assessment of advanced knowledge and complex reasoning abilities. However, evaluating domain-specific LLMs necessitates more than just general and universal capabilities, the desired domain-specific capabilities are equally essential.

Finance is the backbone of modern society, playing a vital role in facilitating economic growth, stability, and prosperity~\cite{shiller2013finance}. However, mastering the intricacies of financial knowledge is challenging for individuals, due to its intricate nature and dynamic environment. Therefore, endowing LLMs with financial knowledge is essential as it can provide significant convenience and insight to humanity. For example, BloombergGPT~\cite{wu2023bloomberggpt} with 50 billion parameters demonstrates superior performance across multiple financial tasks after being trained on a mixed corpus that spans both general and financial domains. FinMA~\cite{xie2023pixiu} is developed by fine-tuning LLaMA~\cite{touvron2023llama} with the financial instruction data to be able to follow instructions for various financial tasks. Similarly, a comprehensive financial evaluation benchmark is also essential for financial LLMs.

\begin{figure}[t]
	\centering
	\includegraphics[width=1.0\linewidth]{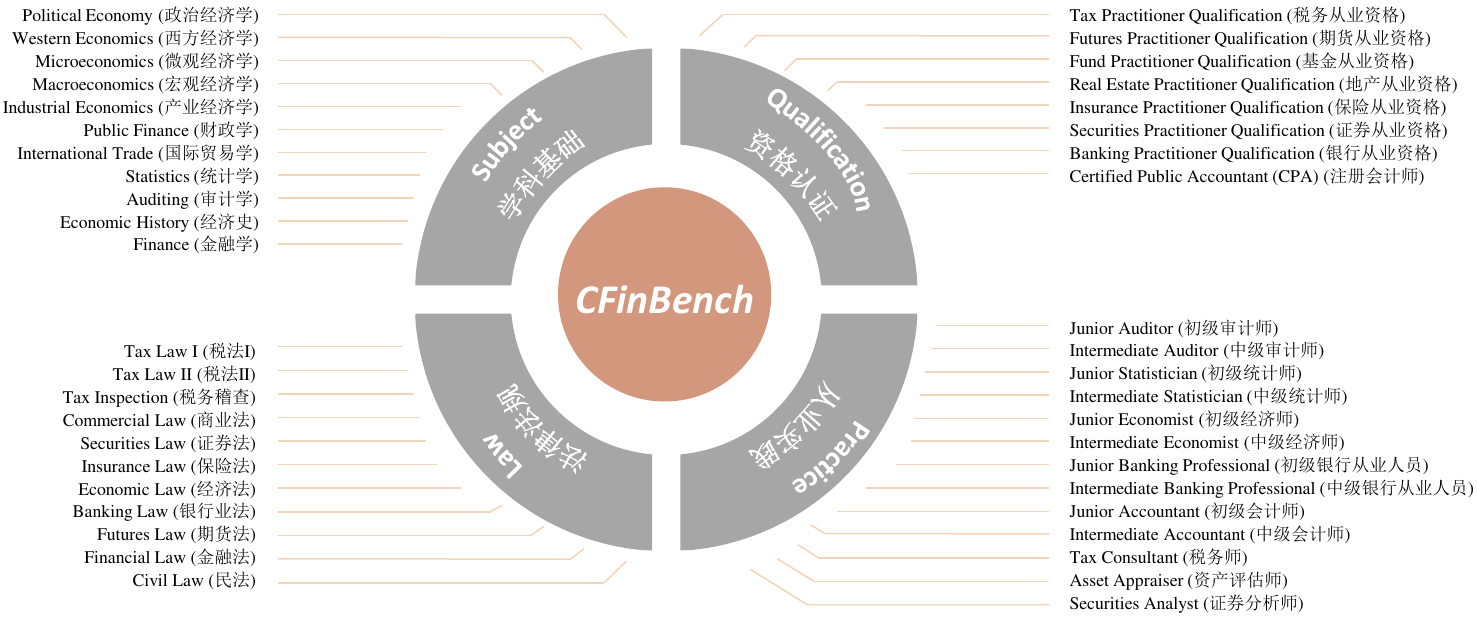}
	\caption{CFinBench comprises 4 first-level categories and 43 second-level categories, which are more align with the career trajectory of financial practitioners.}
	\label{fig0}
	\vspace{-1.0em}
\end{figure}

Several benchmarks have been introduced for better evaluating financial LLMs. FLUE~\cite{shah2022flue} first introduces a financial evaluation benchmark across 5 NLP tasks in English domain, and its successor, FLARE~\cite{xie2023pixiu}, further extends it with financial time-series reasoning task like stock price movements forecasting. In addition to English domain, benchmarks in Chinese are another one of significant importance. BBT-CFLEB~\cite{lu2023bbt} presents the first Chinese financial evaluation benchmark, which includes 6 datasets covering both understanding and generation tasks. FinEval~\cite{zhang2023fineval} builds a collection of 4,661 single-choice questions including 4 first-level categories: finance, economy, accounting, and certificate. However, FinEval is constrained by its limited size, its category coverage is also inadequate for capturing the complexity of real-world financial scenarios. Moreover, the BBT-CFLEB, which targets the basic NLP tasks in finance, struggles to provide sufficient challenges for the increasingly advanced large language models.

In this paper, we present CFinBench: a meticulously crafted, the most comprehensive evaluation benchmark to date, for assessing the capabilities of LLMs on Chinese financial tasks. The design philosophy of our benchmark aligns with the career progression trajectory of financial practitioners, which can be likened to a 'leveling up' in a game. Specifically, it begins with mastering the required foundational knowledge in financial subjects, followed by obtaining the necessary qualified certifications, and subsequently honing skills through practical experience in industry applications. Last but not least, compliance with financial laws and regulations is also a crucial aspect that warrants careful consideration. Formally, we include 4 first-level categories: (1) \emph{Financial Subject}: examining whether LLMs can memorize the foundational knowledge in financial subjects, such as political economy, statistics, macroeconomics, \etc (2) \emph{Financial Qualification}: examining whether LLMs can obtain the necessary qualified certifications for financial practitioners, such as tax practitioner qualification, futures practitioner qualification, fund practitioner qualification, \etc (3) \emph{Financial Practice}: examining whether LLMs can fufill the specific tasks in financial jobs, such as junior banking professional, asset appraiser, junior statistician, \etc (4) \emph{Financial Law}: examining whether LLMs can comply with the financial laws and regulations, such as securities law, insurance law, economic law, \etc CFinBench are primarily sourced from the mock exams freely available on the Internet. To enhance the quality and diversity of the benchmark, and mitigate the problem of data contamination, we perform a series of rigorous data processing pipelines, encompassing data cleaning, internal and external de-duplication, LLM-assisted rephrasing, option shuffling, and multi-round human-in-the-loop cross-validation. CFinBench comprises 99,100 questions spanning 43 second-level categories with 3 question types: single-choice, multiple-choice and judgment. 

We conduct extensive experiments of 50 representative LLMs with various model size on CFinBench. The results show that GPT4 and some Chinese-oriented models like Yi~\cite{young2024yi}, Qwen~\cite{bai2023qwen} and XuanYuan~\cite{zhang2023xuanyuan}, \etc lead the benchmark, with the highest average accuracy being 60.16\%, highlighting the challenge presented by CFinBench and indicating that there is still significant room for improvement for current LLMs in the Chinese financial domain.

\begin{table}[t]
	\setlength{\belowcaptionskip}{1pt}
	\caption{Comparison of the proposed CFinBench with other Chinese-oriented financial benchmarks.}
	\small
	%	 \footnotesize
	\centering
	\setlength{\tabcolsep}{1.2mm}{\begin{tabular}{cccccc}
			\toprule
			\textbf{Benchmark} & \textbf{Month/Year} & \textbf{\#Questions} & \textbf{\#Categories} & \textbf{\#Question Types}& \textbf{Task}\\
			\midrule
			BBT-CFLEB~\cite{lu2023bbt} &2/2023 & 20,416& 6 & 4 & Basic NLP \\
			CGCE~\cite{zhang2023cgce} & 5/2023 & 150& 4 & 1 & QA \\
			CFBenchmark~\cite{lei2023cfbenchmark} & 11/2023 & 3,917& 8 & 3 & Basic NLP \\
			\midrule
			FinEval~\cite{zhang2023fineval} & 8/2023 & 4,661& 34 & 1 & Advanced Knowledge \\
			FinanceIQ~\cite{zhang2023xuanyuan} & 9/2023 & 7,173& 36 & 1 & Advanced Knowledge \\
			Ant-Fin-Eva~\cite{antfineva} & 12/2023 & 8,445& 33 & 1 & Advanced Knowledge \\
			\midrule
			\textbf{CFinBench} & 6/2024 & 99,100& 43 & 3 & Advanced Knowledge \\
			\bottomrule
	\end{tabular}}
	\vspace{-1.0em}
	\label{tab0}
\end{table}

\section{Related Work}
\paragraph{Large Language Models}
The advent of ChatGPT~\cite{chatgpt} marks a significant milestone in natural language processing (NLP), demonstrating the remarkable capabilities of large language models with billions of parameters across a diverse range of tasks. This progress is further amplified by the release of GPT4~\cite{openai2023gpt4}, which exhibits even greater generalization abilities. However, the accompanying strict commercial terms limit the prosperity of the open source community. The seminal contribution in the realm of LLMs is LLaMA~\cite{touvron2023llama,touvron2023llama2,llama3}, an ensemble of publicly-available foundational language models spanning from 7 billion to 70 billion parameters. The up-to-date model, LLaMA3~\cite{llama3}, demonstrates comparable performance to certain proprietary models based on human evaluations. In addition to the English domain orientated LLMs, multilingual models are also thriving. Several notable open-source LLMs, such as Baichuan~\cite{yang2023baichuan}, Qwen~\cite{bai2023qwen}, InternLM~\cite{team2023internlm, cai2024internlm2}, ChatGLM~\cite{zeng2022glm}, \etc have made significant strides. There have also been studies that are dedicated to adapting LLMs to the financial domain. BloombergGPT~\cite{wu2023bloomberggpt}, the first proprietary LLM with 50 billion parameters specialized for the financial domain, is trained with a mixed general and financial corpus. The successors like FinGPT~\cite{yang2023fingpt} and FinMA~\cite{xie2023pixiu} introduce the financial LLMs based on fine-tuning LLaMA, by means of low-rank adaptation or full-parameters. Also, the Chinese-oriented financial LLMs like DISC-FinLLM~\cite{chen2023disc}, CFGPT~\cite{li2023cfgpt}, Xuanyuan~\cite{zhang2023xuanyuan} and YunShan~\cite{wang2023pangu}, have also demonstrated excellent performance across multiple Chinese financial tasks. At the same time, to objectively and quantitatively measure the capabilities of LLMs, a comprehensive and thorough evaluation benchmark is crucial.

\paragraph{Evaluation Benchmarks} 
The rapid advancement LLMs has rendered the conventional evaluation benchmarks~\cite{wang2018glue, xu2020clue, borkan2019nuanced, rajpurkar2018know, rajpurkar2018know, narayan2018don}, typically focused on simple tasks, insufficient. A growing body of research works have recently focused on developing more comprehensive and systematic benchmarks to evaluate the capabilities of LLMs from various perspectives. For example, The MMLU benchmark~\cite{hendrycks2020measuring} evaluates LLMs based on multi-domain and multi-task subjects across STEM, humanities, social sciences, and more. The BIG-bench~\cite{srivastava2022beyond} includes 204 tasks, such as linguistics, math, common-sense reasoning, physics, and beyond. The HELM benchmark~\cite{liang2022holistic} encompasses 42 diverse tasks and employs 7 evaluation metrics, ranging from accuracy to robustness, to assess the performance of LLMs. Despite the flourishing of English benchmark, there have also been efforts in constructing benchmarks for Chinese such as C-Eval~\cite{huang2024c}, CMMLU~\cite{li2023cmmlu}, GAOKAO-Bench~\cite{zhang2023evaluating}, AGIEval~\cite{zhong2023agieval} and Xiezhi~\cite{gu2024xiezhi}, \etc The extensive evaluations on these benchmarks profoundly reveal the advantages and shortcomings of LLMs from multiple capability dimensions. In financial domain, FLUE~\cite{shah2022flue} first introduces a financial evaluation benchmark across 5 NLP tasks in English context, and its successor, FLARE~\cite{xie2023pixiu}, further includes financial time-series reasoning task like stock movement prediction. The FinBen~\cite{xie2024finben} reorganizes 35 public English datasets across 23 financial tasks into three spectrums of difficulty. For the Chinese domain, BBT-CFLEB~\cite{lu2023bbt} presents the first Chinese financial evaluation benchmark, which includes 6 datasets covering both understanding and generation tasks. By integrating FLARE~\cite{xie2023pixiu} and BBT-CFLEB~\cite{lu2023bbt}, ICE-FLARE~\cite{hu2024no} enables the evaluation of bilingual financial tasks. CFBenchmark~\cite{lei2023cfbenchmark} assesses the text processing capabilities across recognition, classification, and generation tasks. CGCE~\cite{zhang2023cgce} incorporates 200 general and 150 finance-specific question-answering questions. The most similar works to ours are FinanceIQ~\cite{zhang2023xuanyuan} and FinEval~\cite{zhang2023fineval}. FinanceIQ~\cite{zhang2023xuanyuan} encompasses 10 primary categories, including economists, securities professionals and actuaries, \etc with a total of 36 subcategories and 7,173 single-choice questions. FinEval~\cite{zhang2023fineval} includes four primary categories, \ie finance, economy, accounting, and certificate, with a total of 34 subcategories and 4,661 single-choice questions. However, these two benchmarks are limited in size. In contrast, the proposed CFinBench is more comprehensive, comprising a more reasonable dimension of financial abilities, with a total of 99,100 questions across 3 question types. The detailed comparison of the CFinBench with other Chinese-oriented financial benchmarks is summarized in Table~\ref{tab0}. It is worth highlighting that we are not a aggregation of existing benchmarks, but rather a significant supplement from the perspective of professional practitioners in terms of advanced knowledge and complex reasoning, which, in conjunction with preceding benchmarks~\cite{shah2022flue,lu2023bbt,zhang2023fineval,lei2023cfbenchmark}, foster the development of financial LLMs.

\section{CFinBench}
In this section, we first introduce the overall design principle of CFinBench, then we elaborate the taxonomy of CFinBench including financial subject, financial qualification, financial practice, and financial law. The detailed procedure of data construction is also presented at the end.

\subsection{Overview}
The motivation of CFinBench is to evaluate the financial knowledge of large language models in the context of Chinese. Inspired by predecessors~\cite{huang2024c,fei2023lawbench,zhang2023fineval}, we also focus on the advanced knowledge and complex reasoning
\begin{figure}[h]
	\centering
	\includegraphics[width=1.0\linewidth]{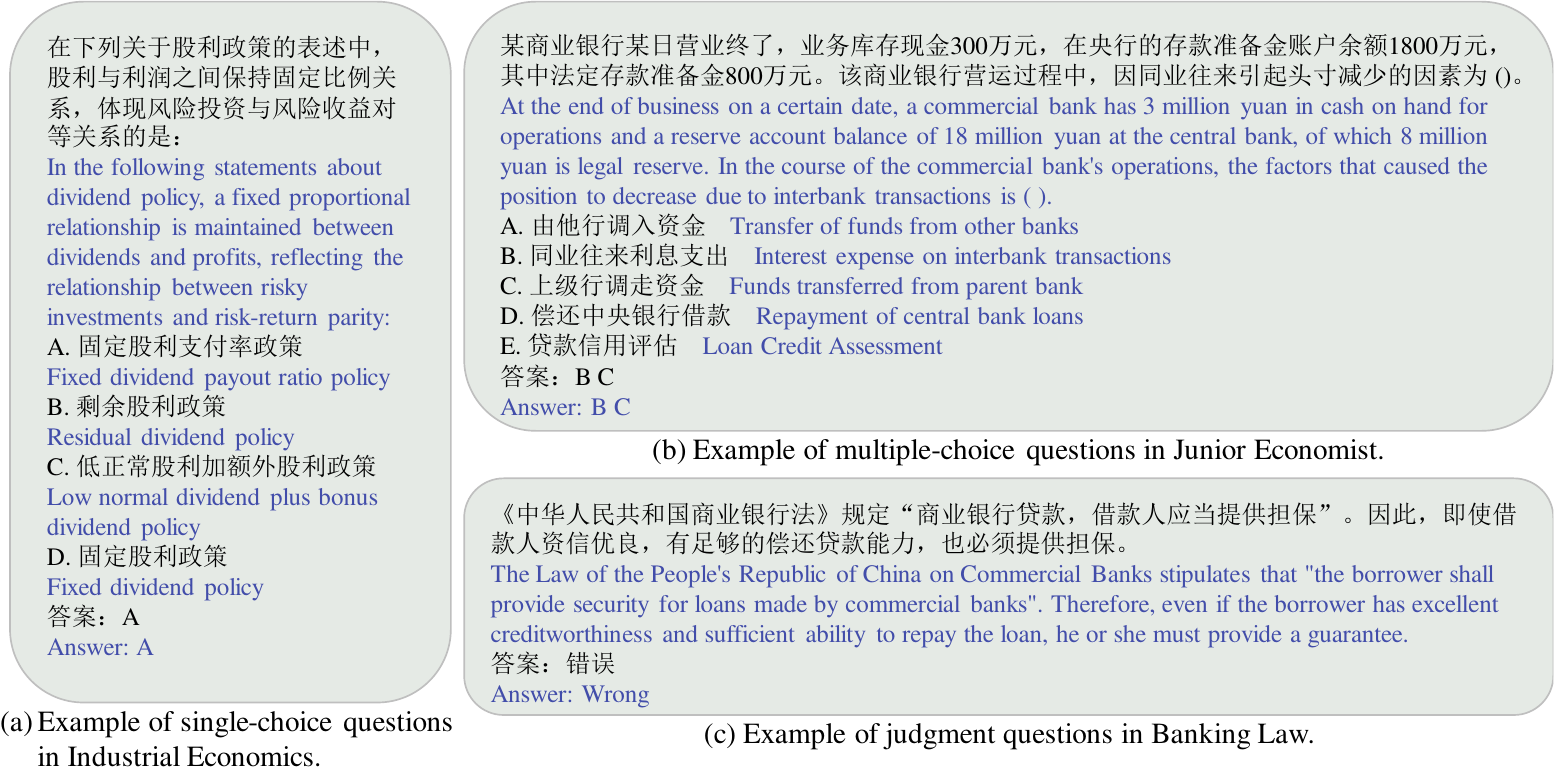}
	\caption{Examples of 3 types of questions in CFinBench. English translations are shown in blue for better readability.}
	\label{fig1}
%	\vspace{-1.0em}
\end{figure}
abilities, which, compared to traditional NLP capabilities, pose a greater challenge to the increasingly advanced LLMs of today. In practice, we build CFinBench based on the real-world examination questions used in China for assessing financial professionals across multiple dimensions. We include 3 question types: single-choice, multiple-choice and judgment, as exemplified in Figure~\ref{fig1}. Compared with the single-choice questions alone in most existing works~\cite{zhang2023fineval, zhang2023xuanyuan, clark2018think, antfineva, huang2024c}, a broader range of question types can more comprehensively assess the capabilities of LLMs. Specifically, for single-choice questions, each question has four options, with only one correct answer. For multiple-choice questions, each question has four or five options, with at least two correct answers. For judgment questions, each question requires a direct judgment of whether the statement is correct or wrong. With prompts, LLMs are expected to answer these questions correctly.

\subsection{Taxonomy} 
In the evaluation of large language models, a diverse array of tasks is often preferred to comprehensively assess their capabilities. A hierarchical evaluation framework enables a more nuanced understanding of the abilities of LLMs. Instead of categorizing the financial tasks solely based on their subjects~\cite{zhang2023fineval, zhang2023xuanyuan}, we thoroughly explore the characteristics of Chinese financial system, and reorganize the financial tasks into more reasonable categories. Specifically, the process starts with acquiring fundamental knowledge in financial subjects, followed by passing essential professional qualifications, and subsequently refining skills through practical experience in industry applications. Additionally, adherence to laws and regulations is a critical aspect that demands careful consideration. In practice, we include 4 first-level categories and 43 second-level categories, which are summarized in Figure~\ref{fig0}.

\begin{itemize}
	\item \textbf{Financial Subject}: The purpose of the financial subject is to test whether LLMs can memorize the essential foundational knowledge in financial subjects. Specifically, 11 subjects are included: Political Economy, Western Economics, Microeconomics, Macroeconomics, Industrial Economics, Public Finance, International Trade, Statistics, Auditing, Economic History, and Finance. These subjects collectively provide a comprehensive framework for understanding the intricacies of economic systems, market structures, and financial institutions. By delving into these areas, individuals can develop a profound understanding of economic theories, policy analyses, market dynamics, and financial instruments, thereby enabling them to make informed decisions, analyze economic trends, and navigate the complexities of global financial markets. In summary, these subjects are crucial in equipping professionals with the foundational financial knowledge.
	
	\item \textbf{Financial Qualification}: The objective of the financial qualification is to examine whether LLMs can obtain necessary qualified certifications for finance professionals. We include 8 qualifications: Tax Practitioner Qualification, Futures Practitioner Qualification, Fund Practitioner Qualification, Real Estate Practitioner Qualification, Insurance Practitioner Qualification, Securities Practitioner Qualification, Banking Practitioner Qualification, and Certified Public Accountant (CPA). These qualifications demonstrate an individual's expertise and proficiency in specific areas of finance, such as taxation, financial markets, investment management, and accounting. By obtaining these qualifications, professionals can enhance their knowledge and skills in areas such as financial analysis, risk management, and financial planning. These qualifications indicate the valuable entry tickets for financial practitioners.
	
	\item \textbf{Financial Practice}: The category of financial practice is to examine whether LLMs can fufill the specific tasks in financial practical jobs. We include 13 titles in practical financial jobs: Junior/Intermediate Auditor, Junior/Intermediate Statistician, Junior/Intermediate Economist, Junior/Intermediate Banking Professional, Junior/Intermediate Accountant, Tax Consultant, Asset Appraiser, and Securities Analyst. These practices involve the application of financial concepts and techniques to real-world problems, requiring professionals to possess a deep understanding of financial markets, instruments, and regulations. By engaging in these practices, professionals can develop expertise in areas like financial reporting, data analysis, economic modeling, risk assessment, and investment analysis. In a nutshell, the ability to accomplish practical financial tasks is one of the key indicators of the competence of financial professionals.
	
	\item \textbf{Financial Law}: The purpose of the financial law is to test whether LLMs can comply with financial laws and regulations. Specifically, it includes 11 exams of laws and regulations: Tax Law I/II, Tax Inspection, Commercial Law, Securities Law, Insurance Law, Economic Law, Banking Law, Futures Law, Financial Law and Civil Law. These laws provide the legal foundation for financial transactions, investments, and operations, and are essential for professionals to understand in order to navigate the financial landscape effectively. Familiarity with these laws enables individuals to appreciate the legal implications of financial decisions, identify potential risks and liabilities, and ensure compliance with regulatory requirements. In summary, proficiency in financial laws can reduce the occurrence of illegal activities and minimize the risk of a 'one-vote veto' in the careers of financial practitioners.
\end{itemize}

\subsection{Data Construction}
\subsubsection{Data Sources} 
Our dataset is primarily composed of mock exams sourced from publicly accessible channels. Notably, some questions of financial qualification and financial practice originate from internal examinations of the financial departments of Chinese companies and are non-publicly available, which are challenging to acquire through web crawling.

\subsubsection{Data Processing} 
\begin{figure}[t]
	\centering
	\includegraphics[width=0.98\linewidth]{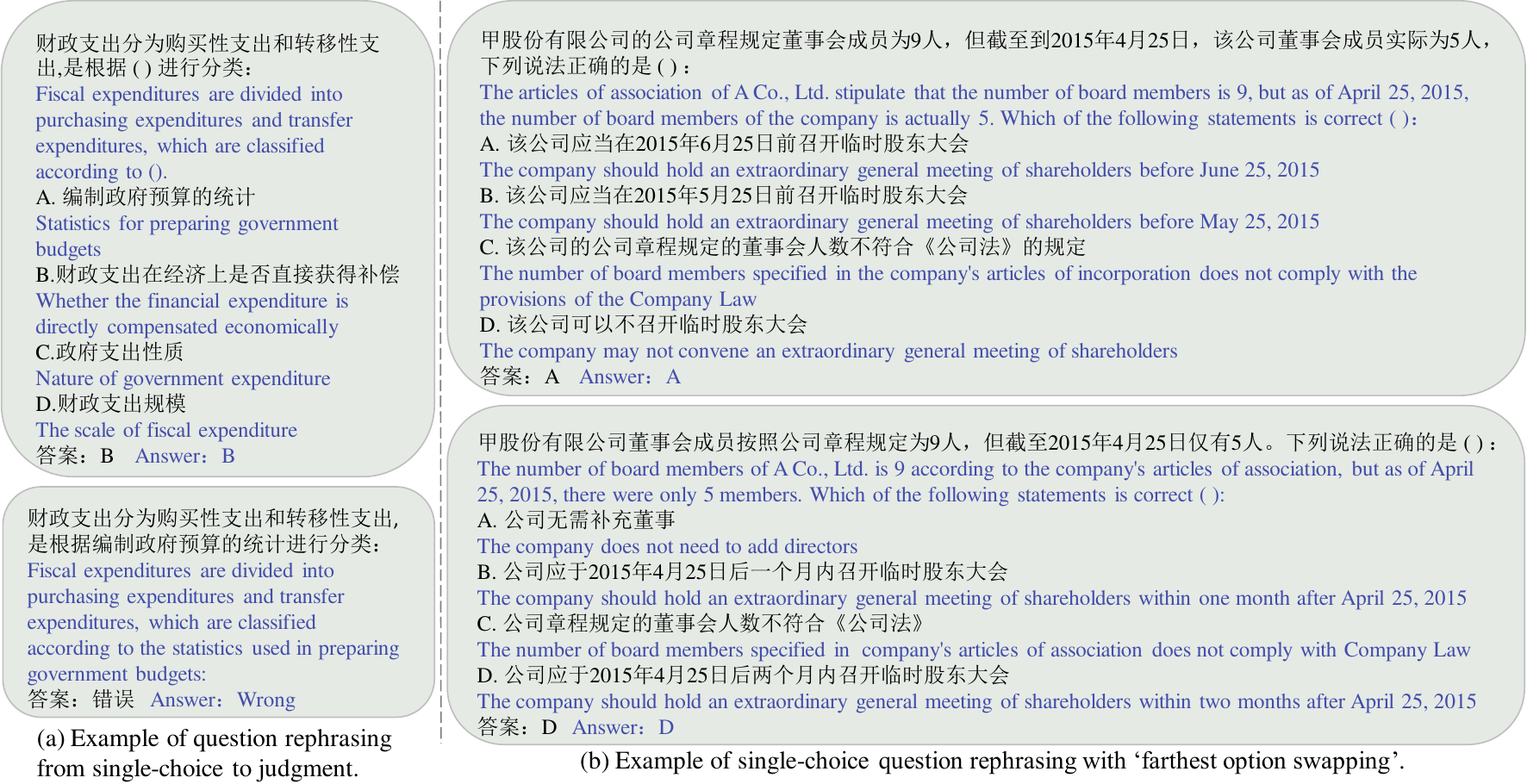}
	\caption{Examples of question rephrasing. English translations are shown in blue for better readability. In each example, the top is the original question, and the bottom is the rephrased question.}
	\label{fig2}
	\vspace{-1.0em}
\end{figure}
The collected data come in various formats, including PDF, EPUB, Microsoft Word documents and web pages. Documents in PDF format and EPUB format are parsed into text using PyMuPDF and EbookLib respectively. We standardized all single-choice questions to have exactly four options, excluding those with fewer options and randomly removing excess wrong options from those with more than four. Similarly, for multiple-choice questions, to maintain uniformity, we only retain questions with four or five options.

Following predecessors~\cite{yuan2021wudaocorpora, penedo2023refinedweb, wei2023skywork, huang2024c}, all the collected questions go through a standard data preprocessing pipeline including cleaning and de-duplication. For data cleaning, we first remove non-Chinese paragraphs with the inexpensive n-gram models like fastText~\cite{joulin2016fasttext}. Then a series of filtering rules and heuristics are performed, such as only keeping lines with valid punctuation, discarding consecutive newlines and whitespace characters, or removing unsemantic and garbled lines. For data de-duplication, we adopt MinHash algorithm~\cite{broder1997resemblance} for internal de-duplication and de-duplication with external public data~\cite{zhang2023fineval, zhang2023xuanyuan, antfineva, openfindata, lu2023bbt}.

To enhance data diversity and mitigate data contamination problem, we also adopt the strategy of question rephrasing based on GPT4~\cite{wang2022self, xu2023wizardlm}. We observe that the collected raw data exhibits a significant class imbalance, with a notable scarcity of judgment questions and a substantial surplus of single-choice questions. To address this issue, we prompt GPT4 to rephrase a portion of the single-choice questions into judgment questions, while maintaining semantic consistency, as exemplified in Figure~\ref{fig2} (a). Furthermore, to mitigate the problem of data contamination, we first randomize the option order~\cite{berglund2023reversal}. In practice, this includes both random shuffle and 'farthest option swapping', where the correct option is exchanged with the incorrect option that is farthest away. Subsequently, we prompt GPT4 to rephrase the questions based on the shuffled options, similarly preserving semantic consistency, as exemplified in Figure~\ref{fig2} (b). Following this,
all the questions undergo several rounds of rigorous human cross-validation. The statistics of the final dataset are summarized in Appendix.

\section{Experiments} 
\subsection{Setup}
\paragraph{Data Split}
We randomly split the questions into a development set, a validation set, and a test set within each second-level category. The development split per category consists of three examples to facilitate few-shot evaluation. A portion of the development examples are also annotated with detailed explanations to enable few-shot chain-of-thought settings~\cite{wei2022chain}, which will be discussed in Appendix. The validation set and test set are divided in a ratio of 2:8, where the validation set is for hyperparameter tuning and the test set is for full evaluation.

\paragraph{Inference Details}
We employ the OpenCompass~\cite{2023opencompass} framework to perform model inference. Specifically, during the generation process, we set both the temperature and the top $p$ to 1.0, and employ greedy decoding. The input token length is limited to 2048, and the output token length is limited to 64, which is sufficient for the questions of choice and judgment. Right truncation is performed for input prompts exceeding the length limitation. All models are inferred in both zero-shot and three-shot settings, which are exemplified in Appendix.

\paragraph{Evaluation Metrics}
We adopt accuracy to measure the match between model prediction and gold answer. Specifically, for single-choice questions, if multiple valid options are predicted by the model, we only select the first option as the final answer predicted by the model. For multiple-choice questions, if any of the options predicted by the model are not among the gold answer, we directly classify it as wrong. Otherwise, we score it based on the number of predicted answers (out of a full score of 1). At last, we calculate the final score for each category based on: $final=0.4 \times single + 0.4 \times multiple + 0.2 \times judgment$.

\subsection{Models}
To give a comprehensive view of the status of LLMs in a Chinese financial context, we evaluate a wide spectrum of large language models, as depicted in Table~\ref{tab1}. Specifically, our experiments cover 46 open-source LLMs from various families, including Llama~\cite{touvron2023llama2,llama3}, Qwen~\cite{bai2023qwen}, ChatGLM~\cite{zeng2022glm,du2022glm}, Baichuan~\cite{yang2023baichuan}, InternLM~\cite{team2023internlm, cai2024internlm2}, Phi~\cite{gunasekar2023textbooks,textbooks2,abdin2024phi}, DeepSeek~\cite{deepseekv2}, XuanYuan~\cite{zhang2023xuanyuan}, FinMA~\cite{xie2023pixiu}, Gemma~\cite{team2024gemma}, TigerBot~\cite{chen2023tigerbot}, Skywork~\cite{wei2023skywork}, Yi~\cite{young2024yi} and Mistral~\cite{jiang2023mistral}. We classify models into different categories according to their size, including greater than 65B, approximately equal to 30B, 10B-20B, 5B-10B, and less than 5B. Considering the legal issues, we only report the results of two publicly recognized API-based LLMs, \ie ChatGPT~\cite{chatgpt} and GPT4~\cite{openai2023gpt4}. In addition, the proprietary finance-specific model YunShan~\cite{wang2023pangu} is included.

\begin{table}[t]
	\centering
	\caption{The 0-shot and 3-shot accuracy (\%) on the test split under the answer-only setting.}
	\scriptsize
	\setlength{\tabcolsep}{2.4mm}\begin{tabular}{lcccccccccc}
		\toprule
		\multirow{2}*{Model} & \multicolumn{2}{c}{Subject}  & \multicolumn{2}{c}{Qualification} & \multicolumn{2}{c}{Practice} & \multicolumn{2}{c}{Law} & \multicolumn{2}{c}{Average} \\
		~ & 0-shot & 3-shot & 0-shot & 3-shot & 0-shot & 3-shot & 0-shot & 3-shot & 0-shot & 3-shot  \\ 
		\midrule
		\multicolumn{11}{l}{\cellcolor[HTML]{EFEFEF}\textit{\textbf{API Call }}}    \\  
		\midrule
		GPT4 & 57.79 & 57.09 & 56.77 & 55.57 & 55.62 & 54.87 & 53.01 & 51.21 & 55.80 & 54.69 \\
		ChatGPT & 39.58 & 40.86 & 43.15 & 42.56 & 40.86 & 40.51 & 38.18 & 38.83 & 40.44 & 40.69 \\
		\midrule
		\multicolumn{11}{l}{\cellcolor[HTML]{EFEFEF}\textit{\textbf{Size > 65B }}}    \\  
		\midrule
		Qwen-72B  & 59.95 & 59.94 & 58.61 & 59.95 & 57.27 & 59.33 & 55.06 & 55.12 & 57.72 & 58.56 \\
		Qwen1.5-72B  & 58.91 & 60.37 & 57.36 & 58.94 & 56.05 & 58.22 & 53.54 & 54.87 & 56.47 & 58.10 \\
		XuanYuan2-70B-Base & 53.56 & 59.96 & 53.36 & 56.27 & 53.55 & 58.05 & 48.27 & 52.48 & 52.19 & 56.69 \\
		XuanYuan-70B-Base & 50.34 & 58.62 & 50.65 & 56.58 & 52.31 & 56.29 & 48.50 & 54.61 & 50.45 & 56.53 \\
		Llama3-70B  & 50.75 & 56.27 & 47.52 & 52.35 & 45.17 & 51.24 & 44.62 & 49.26 & 47.02 & 52.28 \\
		DeepSeek-67B-Base  & 45.76 & 52.03 & 44.61 & 50.57 & 43.95 & 49.16 & 42.87 & 47.01 & 44.30 & 49.69 \\
		Tigerbot-70B-Base & 43.22 & 52.13 & 43.15 & 48.42 & 40.32 & 46.00 & 38.56 & 45.87 & 41.31 & 48.11 \\
		Llama2-70B & 30.03 & 30.26 & 29.88 & 29.06 & 27.47 & 29.75 & 29.68 & 28.26 & 29.27 & 29.33 \\
		\midrule
		\multicolumn{11}{l}{\cellcolor[HTML]{EFEFEF}\textit{\textbf{Size $\approx$ 30B }}}    \\  
		\midrule
		Yi1.5-34B & 58.62 & 61.44 & 58.30 & 60.91 & 57.26 & 59.75 & 55.16 & 58.55 & 57.34 & 60.16 \\
		Qwen1.5-32B & 56.58 & 58.71 & 56.57 & 59.12 & 54.07 & 57.14 & 53.37 & 55.59 & 55.15 & 57.64 \\
		\midrule
		\multicolumn{11}{l}{\cellcolor[HTML]{EFEFEF}\textit{\textbf{10B < Size < 20B }}}    \\  
		\midrule
		Qwen-14B  & 46.07 & 49.82 & 48.68 & 51.35 & 46.35 & 49.47 & 44.82 & 47.59 & 46.48 & 49.56 \\
		InternLM2-20B  & 47.98 & 48.65 & 49.22 & 48.70 & 47.37 & 47.37 & 44.10 & 44.54 & 47.17 & 47.32 \\
		XuanYuan-13B-Base  & 40.87 & 44.32 & 41.64 & 47.68 & 42.30 & 46.56 & 41.73 & 45.75 & 41.64 & 46.08 \\
		InternLM-20B  & 41.74 & 45.15 & 43.41 & 45.25 & 41.11 & 42.33 & 42.05 & 41.39 & 42.08 & 43.53 \\
		Phi3-14B-Instruct & 44.95 & 42.46 & 46.12 & 43.60 & 44.26 & 40.83 & 42.17 & 39.61 & 44.38 & 41.63 \\
		Baichuan2-13B  & 29.16 & 41.13 & 34.25 & 44.19 & 31.27 & 40.63 & 31.44 & 40.05 & 31.53 & 41.50 \\
		Skywork-13B  & 34.66 & 39.24 & 37.78 & 43.88 & 36.22 & 40.90 & 36.39 & 41.38 & 36.26 & 41.35 \\
		Baichuan-13B  & 31.33 & 38.83 & 32.10 & 40.08 & 29.59 & 36.96 & 29.63 & 37.73 & 30.66 & 39.15 \\
		Tigerbot-13B-Base & 32.83 & 34.09 & 35.36 & 39.21 & 33.09 & 35.87 & 33.75 & 35.52 & 33.76 & 36.17 \\
		Llama2-13B  & 30.24 & 32.15 & 31.42 & 35.18 & 29.35 & 31.92 & 29.48 & 34.33 & 30.12 & 33.40 \\
		\midrule
		\multicolumn{11}{l}{\cellcolor[HTML]{EFEFEF}\textit{\textbf{5B < Size < 10B }}}    \\  
		\midrule
		YunShan-7B  & 52.65 & 53.00 & 52.61 & 51.79 & 53.33 & 52.77 & 52.52 & 52.23 & 52.78 & 52.45 \\
		Yi1.5-9B & 47.85 & 50.51 & 50.45 & 51.03 & 48.26 & 49.20 & 46.41 & 47.03 & 48.24 & 49.44 \\
		Qwen1.5-7B  & 46.26 & 49.18 & 47.97 & 50.28 & 46.52 & 47.93 & 44.66 & 46.04 & 46.35 & 48.36 \\
		ChatGLM3-6B-Base & 46.56 & 46.41 & 47.52 & 49.45 & 46.56 & 48.20 & 43.62 & 45.05 & 46.07 & 47.28 \\
		InternLM2-7B & 46.74 & 45.22 & 47.53 & 45.57 & 44.48 & 42.60 & 42.66 & 41.22 & 45.35 & 43.65 \\
		XuanYuan-6B-Base  & 39.99 & 41.65 & 44.30 & 45.87 & 42.70 & 43.91 & 41.70 & 42.81 & 42.17 & 43.56 \\
		Qwen-7B  & 35.34 & 42.19 & 38.36 & 43.75 & 34.98 & 40.41 & 36.01 & 39.90 & 36.17 & 41.56 \\
		ChatGLM2-6B & 40.81 & 40.03 & 43.77 & 43.47 & 41.48 & 40.48 & 40.35 & 40.06 & 41.60 & 41.01 \\
		Baichuan2-7B  & 30.22 & 37.23 & 35.56 & 41.35 & 31.30 & 37.33 & 29.59 & 37.49 & 31.67 & 38.35 \\
		Llama3-8B  & 25.71 & 37.02 & 29.01 & 40.00 & 25.66 & 38.18 & 26.05 & 36.82 & 26.61 & 38.01 \\
		Mistral-7B & 29.46 & 35.63 & 29.11 & 37.56 & 28.75 & 35.87 & 28.39 & 34.34 & 28.93 & 35.85 \\
		InternLM-7B  & 31.93 & 34.62 & 38.81 & 38.55 & 34.77 & 34.19 & 32.06 & 35.04 & 34.39 & 35.60 \\
		ChatGLM-6B & 36.80 & 34.78 & 36.40 & 36.65 & 36.33 & 35.20 & 35.32 & 34.93 & 36.21 & 35.39 \\
		Gemma-7B & 35.22 & 33.51 & 38.28 & 37.07 & 36.64 & 32.53 & 37.12 & 34.16 & 36.82 & 34.32 \\
		Baichuan-7B  & 27.09 & 31.43 & 29.70 & 34.19 & 26.43 & 30.34 & 27.31 & 31.70 & 27.63 & 31.92 \\
		Tigerbot-7B-Base & 29.70 & 31.19 & 30.10 & 34.40 & 29.16 & 30.44 & 30.31 & 31.28 & 29.82 & 31.83 \\
		Llama2-7B & 27.59 & 29.60 & 29.62 & 33.55 & 27.66 & 30.09 & 28.45 & 31.36 & 28.33 & 31.15 \\
		FinMA-7B  & 23.71 & 22.74 & 24.92 & 25.86 & 22.71 & 20.71 & 22.34 & 23.52 & 23.42 & 23.21 \\
		\midrule
		\multicolumn{11}{l}{\cellcolor[HTML]{EFEFEF}\textit{\textbf{Size < 5B }}}    \\  
		\midrule
		Qwen1.5-4B  & 39.63 & 43.29 & 44.67 & 45.48 & 41.02 & 42.50 & 40.68 & 42.10 & 41.50 & 43.34 \\
		YunShan-1.5B  & 35.06 & 37.36 & 39.19 & 41.61 & 38.17 & 39.95 & 38.08 & 40.32 & 37.63 & 39.81 \\
		Phi3-3.8B-Instruct & 34.15 & 36.98 & 36.22 & 40.66 & 35.74 & 37.86 & 34.17 & 39.28 & 35.07 & 38.70 \\
		Qwen1.5-1.8B  & 33.92 & 35.81 & 38.18 & 41.13 & 33.96 & 36.38 & 36.34 & 37.43 & 35.60 & 37.69 \\
		Qwen-1.8B  & 29.56 & 32.25 & 34.53 & 36.40 & 29.17 & 32.69 & 30.98 & 34.27 & 31.06 & 33.90 \\
		Qwen1.5-0.5B  & 31.20 & 30.83 & 34.70 & 36.85 & 32.09 & 31.10 & 34.00 & 34.68 & 33.00 & 33.37 \\
		InternLM2-1.8B  & 29.59 & 33.86 & 35.04 & 33.86 & 29.58 & 31.49 & 32.40 & 32.66 & 31.65 & 32.97 \\
		Gemma-2B  & 24.95 & 27.36 & 26.24 & 29.47 & 22.47 & 25.92 & 24.48 & 29.93 & 24.54 & 28.17 \\
		Phi2-2.7B & 23.07 & 25.63 & 27.04 & 30.94 & 23.13 & 27.64 & 26.38 & 28.12 & 24.91 & 28.08 \\
		Phi1.5-1.3B & 15.35 & 23.40 & 18.79 & 26.73 & 14.68 & 23.27 & 16.64 & 27.02 & 16.37 & 25.11 \\
		\bottomrule
	\end{tabular}
	\vspace{-1.0em}
	\label{tab1}
\end{table}

\subsection{Results}
In Table~\ref{tab1}, we report the 0-shot and 3-shot accuracy of each first-level category on the test split under the answer-only setting. As can be seen, the Chinese-oriented Yi1.5-34B~\cite{young2024yi} lead the benchmark, with a mean accuracy just reaching 60.16\%, highlighting the challenge presented by CFinBench.  In addition, the accuracy of some other Chinese-oriented models such as Qwen-72B~\cite{bai2023qwen}, Qwen1.5-72B~\cite{bai2023qwen}, XuanYuan2-70B-Base~\cite{zhang2023xuanyuan} and XuanYuan-70B-Base~\cite{zhang2023xuanyuan} also exceed 56\%. The accuracy of GPT4~\cite{openai2023gpt4} is around 55\%, which also shows obvious advantage compared to other models. In the size range of 10B-20B, Qwen-14B~\cite{bai2023qwen}, InternLM2-20B~\cite{cai2024internlm2} and XuanYuan-13B-Base~\cite{zhang2023xuanyuan} are in the lead, with accuracy exceeding 46\%. Notably, in the size range of 5B-10B, the proprietary Chinese finance-specific model YunShan-7B~\cite{wang2023pangu} is in the absolute leading position, with an accuracy of over 52\%, which is even higher than the accuracy of some 70B models. Also, Yi1.5-9B~\cite{young2024yi}, Qwen1.5-7B~\cite{bai2023qwen} and ChatGLM3-6B-Base~\cite{zeng2022glm,du2022glm} also achieve the accuracy over 45\%. During the size range of less than 5B, Qwen1.5-4B~\cite{bai2023qwen} is the only model that achieves an accuracy of more than 40\%. The other models like YunShan-1.5B~\cite{wang2023pangu}, Phi3-3.8B-Instruct~\cite{abdin2024phi} and Qwen1.5-1.8B~\cite{bai2023qwen} also achieve good performance with an accuracy of more than 35\%. In conclusion, there is still significant room for improvement for current LLMs in the Chinese financial domain.

\subsection{Analysis}
\paragraph{Few-shot examples are helpful in most cases.} 
As we can see from Table~\ref{tab1}, the performance of most models demonstrates improvement when some examples are provided. However, in the case of InternLM2-7B and ChatGLM-6B, \etc the zero-shot setting outperforms the few-shot setting. We guess that this is because these models have acquired the ability to fully understand the questions without the need for examples during the pre-training or fine-tuning. The introduced examples may mismatch with their training methodology, which leads to the decrease in accuracy~\cite{gu2024xiezhi,li2023cmmlu}.

\paragraph{Scaling up the model size usually results in better performance.}
In Table~\ref{tab1}, as the model size increases, the accuracy of Qwen1.5-0.5B, Qwen1.5-1.8B, Qwen1.5-4B and Qwen1.5-7B increases accordingly, \ie 33.37\%, 37.69\%, 43.34\% and 48.36\% respectively at the 3-shot setting. Similarly, the accuracy of Yi1.5-34B is increased by 10.72\% over Yi1.5-9B at the 3-shot setting. However, this does not mean that increasing the model size will definitely improve the performance. 

\paragraph{Domain specific pre-training and fine-tuning are helpful.}
Impressively, two finance-specific models XuanYuan~\cite{zhang2023xuanyuan} and YunShan~\cite{wang2023pangu} achieve very competitive accuracy. This can be attributed to the fact that both models are mixed with high-quality financial corpus during pre-training and fine-tuning, resulting in better grasp of financial knowledge. We think the same is true for the general base models such as Yi~\cite{young2024yi} and Qwen~\cite{bai2023qwen} that perform well in Table~\ref{tab1}.

\paragraph{Results on the validation split.}
Since we do not publicly release the labels for the test split, we provide the average accuracy on the validation split as a reference for developers. The results of 0-shot and 3-shot on the validation split and test split under answer-only setting are reported in Table~\ref{tab2}. We can observe that the difference between the results of the same model on validation split and test split is very small. For example, the 0-shot accuracy of GPT4 on validation split and test split differ only by 0.14\%, suggesting that developers can use the accuracy on validation split as a good indicator to iterate their models faster.

\paragraph{The performance of chat models.}
To better engage in natural conversation, the chat version are often derived from base model by alignment techniques~\cite{ouyang2022training}, such as supervised finetuning (SFT) and reinforcement learning from human feedback (RLHF). As observed in Table~\ref{tab3}, the accuracy of some models' chat version is improved when compared to the base version, such as Qwen1.5-32B, InternLM2-20B, Baichuan2-13B, \etc At the same time, the accuracy of some models' chat version have declined, such as Yi1.5-34B, ChatGLM3-6B, and Qwen1.5-1.8B. The varying alignment strategies of the models lead to different results.

\begin{table}[t]
	\begin{minipage}{0.49\linewidth}	
		\captionof{table}{The 0/3-shot average accuracy (\%) on the validation split and test split.}
		\makeatletter\def\@captype{table}
		\small
		\centering
		\setlength{\tabcolsep}{1.1mm}{\begin{tabular}{lcccc}
		\toprule
		\multirow{2}*{Model} & \multicolumn{2}{c}{Val} & \multicolumn{2}{c}{Test} \\
        ~ & 0-shot & 3-shot & 0-shot & 3-shot  \\ 
        \midrule
		GPT4  & 55.66 & 54.63 & 55.80 & 54.69 \\
		Qwen1.5-72B  & 56.33 & 58.17 & 56.47 & 58.10 \\
		Yi1.5-34B & 57.71 & 59.72 & 57.34 & 60.16 \\
		Qwen1.5-32B & 54.94 & 57.85 & 55.15 & 57.64 \\
		InternLM2-20B & 47.17 & 48.37 & 47.17 & 47.32 \\
		Baichuan2-13B & 32.66 & 42.16 & 31.53 & 41.50 \\
		Llama3-8B  & 26.72 & 37.91 & 26.61 & 38.01 \\
		ChatGLM3-6B-Base & 47.34 & 47.90 & 46.07 & 47.28 \\
		\bottomrule
		\end{tabular}}
		\label{tab2}
	\end{minipage}
	\hspace{2.5mm}
	\begin{minipage}{0.49\linewidth}	
		\captionof{table}{The 0/3-shot average accuracy (\%) of base model and chat model on the test split.}
		\makeatletter\def\@captype{table}
		\small
		\centering
		\setlength{\tabcolsep}{1.4mm}{\begin{tabular}{lcccc}
		\toprule
		\multirow{2}*{Model} & \multicolumn{2}{c}{Base} & \multicolumn{2}{c}{Chat} \\
		~ & 0-shot & 3-shot & 0-shot & 3-shot  \\ 
		\midrule
		Yi1.5-34B & 57.34 & 60.16 & 58.99 & 57.48 \\
		Qwen1.5-32B & 55.15 & 57.64 & 59.87 & 58.80 \\
		InternLM2-20B & 47.17 & 47.32 & 48.19 & 45.49 \\
		Baichuan2-13B & 31.53 & 41.50 & 40.74 & 44.60 \\
		Llama3-8B  & 26.61 & 38.01 & 42.04 & 41.73 \\
		ChatGLM3-6B & 46.07 & 47.28 & 30.79 & 27.27 \\
		Gemma-2B  & 24.54 & 28.17 & 34.38 & 33.72 \\
		Qwen1.5-1.8B & 35.60 & 37.69 & 37.50 & 35.78 \\
		\bottomrule
		\end{tabular}}
		\label{tab3}
	\end{minipage}
	\vspace{-1.0em}
\end{table}

\section{Conclusion}
In this paper, we present CFinBench, the most comprehensive evaluation benchmark to date, for assessing the financial domain knowledge of LLMs under Chinese context. We improve the quality and diversity of the data and mitigate the issue of data contamination through a series of processes, including data cleaning, internal and external de-duplication, LLM-assisted question rephrasing, option shuffling, and multiple rounds of human cross-validation.  Four first-level categories are included in CFinBench: financial subject, financial qualification, financial practice, and financial law, which are more align with the career trajectory of financial practitioners. The CFinBench comprises 99,100 questions spanning 43 second-level categories with 3 question types: single-choice, multiple-choice and judgment. We conduct extensive evaluations of 50 representative LLMs with various model size on CFinBench. The results show that there is still significant room for improvement for current LLMs in the Chinese financial domain. 

{\small
	\bibliographystyle{plain}
	\bibliography{ref}
}

%%%%%%%%%%%%%%%%%%%%%%%%%%%%%%%%%%%%%%%%%%%%%%%%%%%%%%%%%%%%
\section*{Checklist}
\begin{enumerate}

\item For all authors...
\begin{enumerate}
  \item Do the main claims made in the abstract and introduction accurately reflect the paper's contributions and scope?
    \answerYes{}
  \item Did you describe the limitations of your work?
    \answerYes{We describe our limitations in Appendix.}
  \item Did you discuss any potential negative societal impacts of your work?
    \answerYes{We discuss the potential negative societal impacts in Appendix.}
  \item Have you read the ethics review guidelines and ensured that your paper conforms to them?
    \answerYes{}
\end{enumerate}

\item If you are including theoretical results...
\begin{enumerate}
  \item Did you state the full set of assumptions of all theoretical results?
    \answerNA{}
	\item Did you include complete proofs of all theoretical results?
    \answerNA{}
\end{enumerate}

\item If you ran experiments (e.g. for benchmarks)...
\begin{enumerate}
  \item Did you include the code, data, and instructions needed to reproduce the main experimental results (either in the supplemental material or as a URL)?
    \answerYes{The dataset and code are available at \url{https://cfinbench.github.io/} }
  \item Did you specify all the training details (e.g., data splits, hyperparameters, how they were chosen)?
    \answerNA{}
	\item Did you report error bars (e.g., with respect to the random seed after running experiments multiple times)?
    \answerNA{}
	\item Did you include the total amount of compute and the type of resources used (e.g., type of GPUs, internal cluster, or cloud provider)?
    \answerNA{}
\end{enumerate}

\item If you are using existing assets (e.g., code, data, models) or curating/releasing new assets...
\begin{enumerate}
  \item If your work uses existing assets, did you cite the creators?
    \answerYes{}
  \item Did you mention the license of the assets?
    \answerYes{Our evaluation code is developed based on OpenCompass~\cite{2023opencompass}, which uses the license of Apache-2.0.}
  \item Did you include any new assets either in the supplemental material or as a URL?
    \answerNo{}
  \item Did you discuss whether and how consent was obtained from people whose data you're using/curating?
    \answerNA{}
  \item Did you discuss whether the data you are using/curating contains personally identifiable information or offensive content?
    \answerNA{}
\end{enumerate}

\item If you used crowdsourcing or conducted research with human subjects...
\begin{enumerate}
  \item Did you include the full text of instructions given to participants and screenshots, if applicable?
    \answerNA{}
  \item Did you describe any potential participant risks, with links to Institutional Review Board (IRB) approvals, if applicable?
    \answerNA{}
  \item Did you include the estimated hourly wage paid to participants and the total amount spent on participant compensation?
    \answerNA{}
\end{enumerate}

\end{enumerate}

%%%%%%%%%%%%%%%%%%%%%%%%%%%%%%%%%%%%%%%%%%%%%%%%%%%%%%%%%%%%

\appendix
\section{Appendix}
\subsection{More experiments}
\paragraph{Results on chain-of-thought setting.}
To further explore the models’ reasoning capabilities, in addition to the answer-only (AO) setting, we also perform some experiments on the chain-of-thought (COT) setting~\cite{kojima2022large, wei2022chain}. Evaluation on COT setting requires the model to generate explanations for a given question and then give the final answer based on the generated explanations. Specifically, we obtain the explanations examples only for multiple-choice questions manually by professional financial practitioners. The experimental results are reported in Table~\ref{tab1}.

\begin{table}[h]
	\centering
	\caption{The 0-shot and 3-shot average accuracy (\%) of multiple-choice questions on the test split under the answer-only (AO) setting and chain-of-thought (COT) setting. }
	\small
	\setlength{\tabcolsep}{2.4mm}\begin{tabular}{lcccccc}
		\toprule
		~ & \multicolumn{2}{c}{GPT4}  & \multicolumn{2}{c}{Qwen-14B} & \multicolumn{2}{c}{ChatGLM3-6B-Base} \\
		~ & 0-shot & 3-shot & 0-shot & 3-shot & 0-shot & 3-shot \\ 
		\midrule
		AO & 47.81 & 47.39 & 36.51 & 40.66 & 37.27 & 37.58  \\
		COT & 47.95 & 47.62 & 27.65 & 29.96 & 29.48 & 31.11  \\
		\bottomrule
	\end{tabular}
	\label{appendix_tab1}
\end{table}

As observed in Table~\ref{appendix_tab1}, the models achieve comparable or lower average accuracy than in the answer-only setting. This suggests that COT prompting does not necessarily improve results, which is also evidenced in other benchmarks like FinEval~\cite{zhang2023fineval} and C-Eval~\cite{huang2024c}, \etc

\begin{table}[h]
	\centering
	\caption{The few-shot average accuracy (\%) of CFinBench and FinEval under the answer-only setting. }
	\small
	\setlength{\tabcolsep}{0.95mm}\begin{tabular}{lccccccccc}
		\toprule
		~ & GPT4  & Yi1.5-34B & Qwen1.5-72B   & Qwen1.5-32B &Qwen1.5-7B & ChatGLM3-6B &InternLM2-7B\\
		\midrule
		CFinBench & 54.69 & 60.16 & 58.56  & 57.64  &48.36 &47.28 & 43.65\\
		FinEval & 68.60 & 86.79 & 83.93   & 84.36 &72.55 & 63.08 &62.03\\
		\bottomrule
	\end{tabular}
	\label{appendix_tab2}
\end{table}

\paragraph{Comparison with other similar benchmark.} FinEval~\cite{zhang2023fineval} is another representative benchmark for evaluating the Chinese financial advanced knowledge of LLMs. In Table~\ref{appendix_tab2}, we report the few-shot accuracy on CFinBench and FinEval with various LLMs. It can be seen that the accuracy of the highest Yi1.5-34B reaches nearly 87\%, while ours is around 60\%. Likewise, the lowest InternLM2-7B accuracy is still over 62\%, while ours is only around 43\%. This all suggests that CFinBench is more challenging and better able to distinguish the performance of different models.

\paragraph{More results on chat models.}
To better engage in natural conversation, the chat version are often derived from the base model by alignment techniques~\cite{ouyang2022training}, such as supervised finetuning (SFT) and reinforcement learning from human feedback (RLHF). In Table~\ref{appendix_tab3}, we report more results of the representative chat models.

\begin{table}[t]
	\centering
	\caption{The 0-shot and 3-shot accuracy (\%) on the test split under the answer-only setting. "Average" column indicates the average accuracy over all the first-level categories.}
	\scriptsize
	\setlength{\tabcolsep}{2.4mm}\begin{tabular}{lcccccccccc}
		\toprule
		\multirow{2}*{Model} & \multicolumn{2}{c}{Subject}  & \multicolumn{2}{c}{Qualification} & \multicolumn{2}{c}{Practice} & \multicolumn{2}{c}{Law} & \multicolumn{2}{c}{Average} \\
		~ & 0-shot & 3-shot & 0-shot & 3-shot & 0-shot & 3-shot & 0-shot & 3-shot & 0-shot & 3-shot  \\ 
		\midrule
		\multicolumn{11}{l}{\cellcolor[HTML]{EFEFEF}\textit{\textbf{Size > 65B }}}    \\  
		\midrule
		Qwen1.5-72B  & 62.24 & 61.52 & 61.50 & 61.76 & 60.87 & 60.88 & 59.03 & 59.75 & 60.91 & 60.98 \\
		XuanYuan2-70B & 50.82 & 55.42 & 51.42 & 52.03 & 51.39 & 52.35 & 48.05 & 52.42 & 50.42 & 53.06 \\
		Llama3-70B  & 51.98 & 53.76 & 53.04 & 52.90 & 51.54 & 51.47 & 48.75 & 50.98 & 51.33 & 52.28 \\
		DeepSeek-67B  & 48.64 & 52.88 & 49.74 & 50.85 & 49.03 & 52.65 & 44.76 & 50.25 & 48.04 & 51.66 \\
		Llama2-70B & 25.78 & 30.32 & 27.96 & 31.32 & 25.02 & 29.02 & 26.04 & 32.84 & 26.20 & 30.88 \\
		\midrule
		\multicolumn{11}{l}{\cellcolor[HTML]{EFEFEF}\textit{\textbf{Size $\approx$ 30B }}}    \\  
		\midrule
		Qwen1.5-32B & 59.79 & 59.37 & 61.22 & 60.00 & 59.65 & 57.91 & 58.80 & 57.90 & 59.87 & 58.80 \\
		Yi1.5-34B & 60.96 & 58.10 & 59.49 & 57.96 & 58.69 & 57.19 & 56.91 & 56.65 & 58.99 & 57.48 \\
		\midrule
		\multicolumn{11}{l}{\cellcolor[HTML]{EFEFEF}\textit{\textbf{10B < Size < 20B }}}    \\  
		\midrule
		Qwen-14B  & 48.74 & 49.35 & 49.66 & 49.81 & 47.82 & 47.83 & 45.66 & 45.9 & 47.97 & 48.22 \\
		XuanYuan-13B  & 43.91 & 45.49 & 47.13 & 49.76 & 45.55 & 48.49 & 44.24 & 47.27 & 45.21 & 47.75 \\
		InternLM2-20B  & 49.78 & 46.06 & 49.46 & 47.95 & 48.22 & 43.29 & 45.29 & 44.67 & 48.19 & 45.49 \\
		InternLM-20B  & 43.22 & 44.89 & 45.55 & 46.88 & 43.75 & 45.82 & 42.05 & 43.43 & 43.64 & 45.26 \\
		Baichuan2-13B  & 41.01 & 44.68 & 42.00 & 45.80 & 40.23 & 44.61 & 39.72 & 43.31 & 40.74 & 44.60 \\
		Tigerbot-13B & 38.55 & 37.26 & 40.29 & 42.32 & 37.97 & 38.76 & 37.73 & 37.39 & 38.64 & 38.93 \\
		Baichuan-13B  & 33.56 & 39.28 & 39.54 & 38.40 & 34.49 & 38.41 & 35.80 & 36.29 & 35.85 & 38.10 \\
		\midrule
		\multicolumn{11}{l}{\cellcolor[HTML]{EFEFEF}\textit{\textbf{5B < Size < 10B }}}    \\  
		\midrule
		Yi1.5-9B & 55.32 & 54.16 & 55.48 & 53.93 & 53.77 & 52.79 & 51.56 & 51.03 & 54.03 & 52.98 \\
		Qwen1.5-7B & 48.32 & 49.98 & 50.45 & 50.12 & 48.34 & 48.15 & 47.26 & 46.95 & 48.59 & 48.80 \\
		InternLM2-7B & 44.93 & 45.73 & 46.67 & 48.04 & 45.51 & 45.99 & 43.85 & 44.49 & 45.24 & 46.06 \\
		XuanYuan-6B  & 44.18 & 42.62 & 45.17 & 46.29 & 45.01 & 44.58 & 42.84 & 41.92 & 44.30 & 43.85 \\
		Llama3-8B  & 41.34 & 41.30 & 43.49 & 42.63 & 42.51 & 41.54 & 40.83 & 41.44 & 42.04 & 41.73 \\
		InternLM-7B  & 38.07 & 41.09 & 42.85 & 44.21 & 40.70 & 41.53 & 38.33 & 39.97 & 39.99 & 41.70 \\
		Mistral-7B & 39.07 & 39.44 & 40.77 & 43.38 & 40.04 & 41.83 & 38.69 & 40.63 & 39.64 & 41.32 \\
		Baichuan2-7B  & 40.24 & 40.13 & 42.59 & 41.81 & 42.77 & 41.58 & 40.68 & 39.88 & 41.57 & 40.85 \\
		Tigerbot-7B & 34.11 & 34.53 & 37.45 & 35.03 & 35.48 & 33.74 & 35.42 & 32.79 & 35.62 & 34.02 \\
		ChatGLM3-6B & 28.68 & 26.14 & 33.53 & 27.66 & 31.08 & 27.64 & 29.88 & 27.63 & 30.79 & 27.27 \\
		\midrule
		\multicolumn{11}{l}{\cellcolor[HTML]{EFEFEF}\textit{\textbf{Size < 5B }}}    \\  
		\midrule
		Qwen1.5-4B  & 43.03 & 43.19 & 46.53 & 45.83 & 44.77 & 43.50 & 42.50 & 42.37 & 44.21 & 43.72 \\
		Qwen-1.8B  & 36.18 & 35.94 & 36.35 & 37.47 & 34.87 & 38.75 & 36.55 & 36.61 & 35.99 & 37.19 \\
		InternLM2-1.8B  & 34.55 & 34.72 & 40.26 & 39.52 & 37.25 & 37.05 & 36.86 & 37.24 & 37.23 & 37.13 \\
		Qwen1.5-1.8B  & 36.23 & 35.18 & 39.51 & 37.69 & 37.15 & 35.53 & 37.12 & 34.71 & 37.50 & 35.78 \\
		Gemma-2B  & 32.77 & 32.26 & 35.53 & 32.96 & 34.60 & 34.57 & 34.60 & 35.09 & 34.38 & 33.72 \\
		Qwen1.5-0.5B  & 34.80 & 31.62 & 35.63 & 30.56 & 35.58 & 29.94 & 36.18 & 32.29 & 35.55 & 31.10 \\
		\bottomrule
	\end{tabular}
	\label{appendix_tab3}
\end{table}

\subsection{Prompt examples}
\begin{figure}[t]
	\centering
	\includegraphics[width=0.98\linewidth]{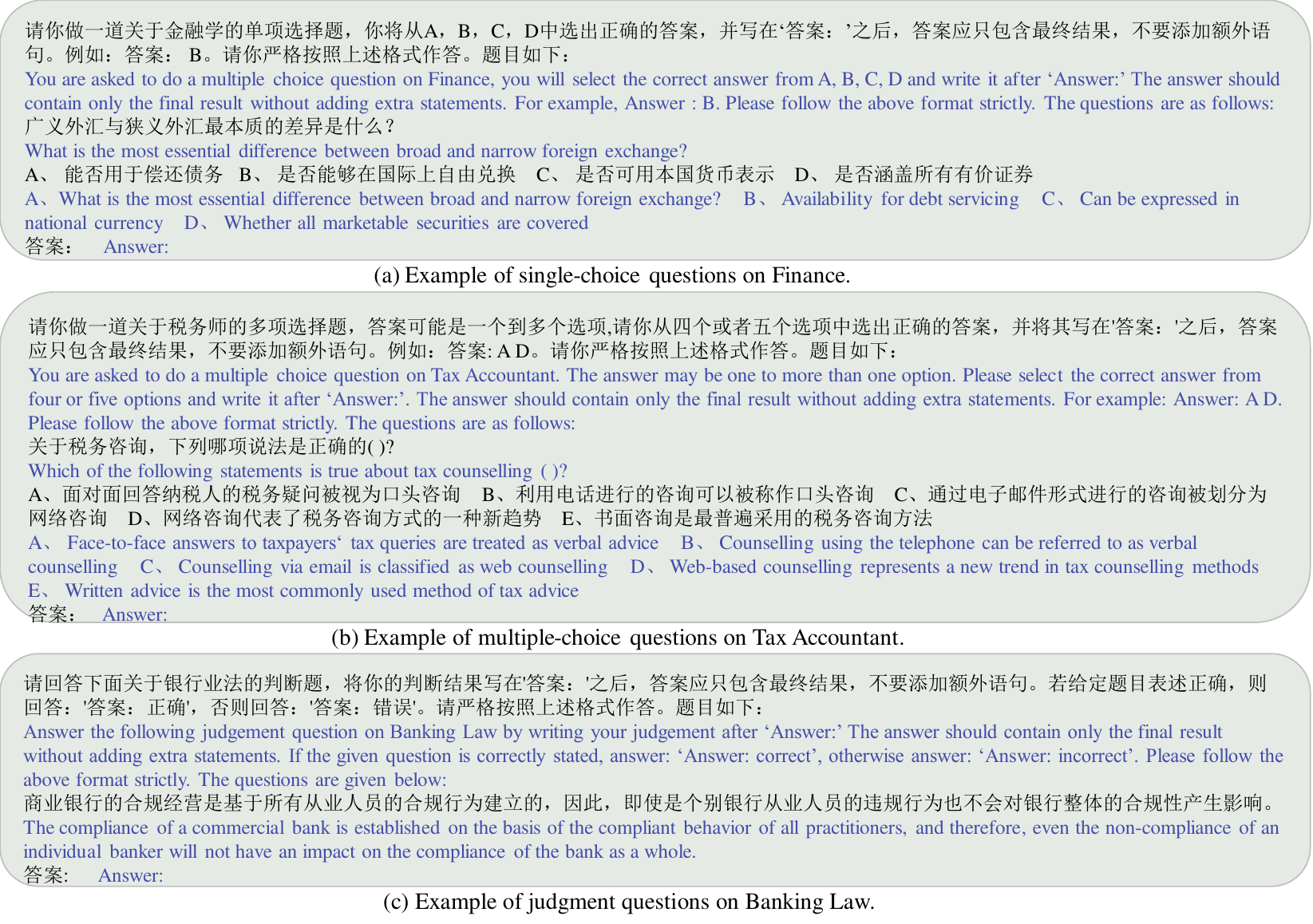}
	\caption{Examples of zero-shot prompts in answer-only setting. English translations are shown in blue for better readability.}
	\label{appendix_fig1}
\end{figure}

\begin{figure}[t]
	\centering
	\includegraphics[width=0.98\linewidth]{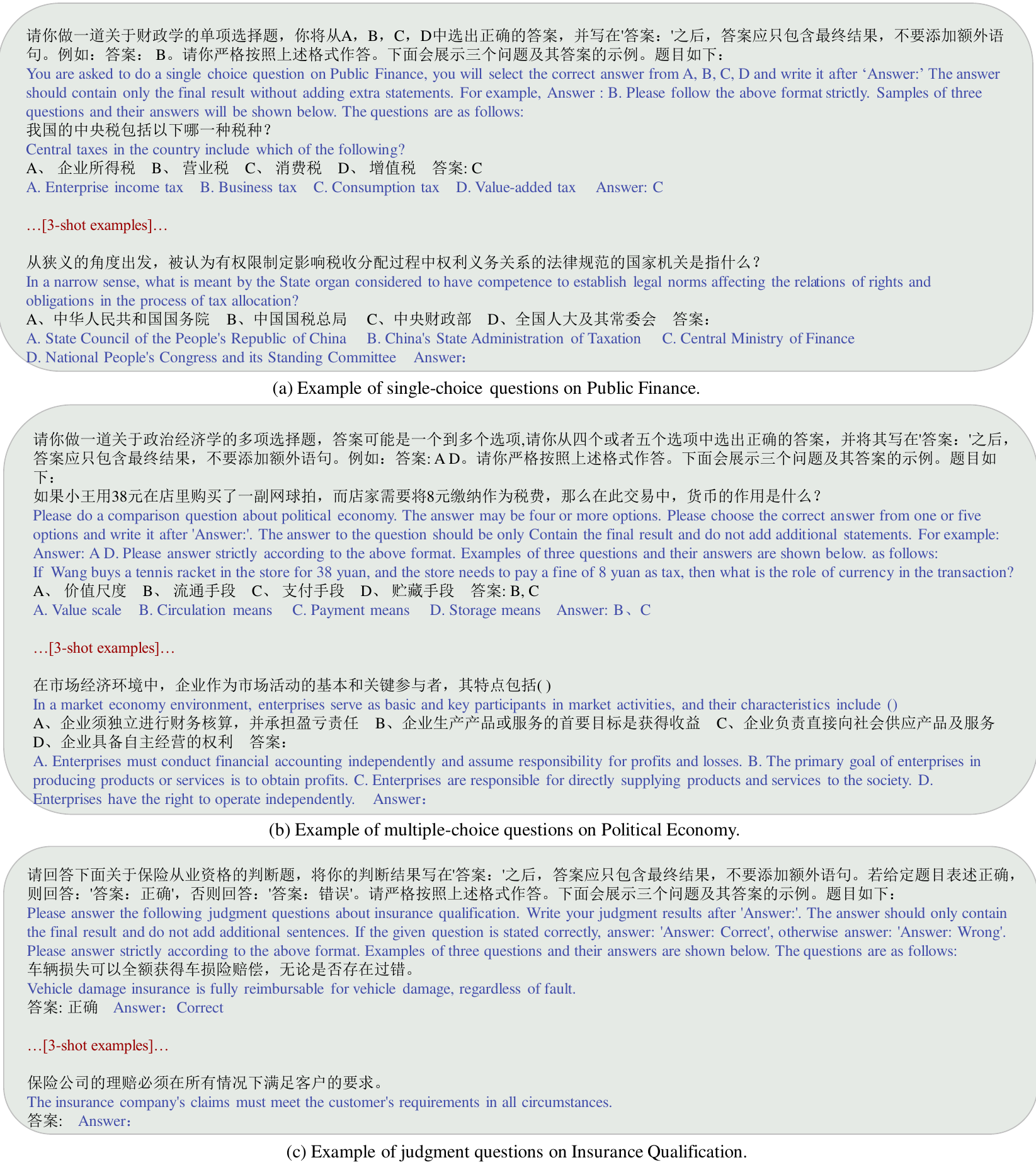}
	\caption{Examples of few-shot prompts in answer-only setting. English translations are shown in blue for better readability.}
	\label{appendix_fig2}
\end{figure}

\begin{figure}[t]
	\centering
	\includegraphics[width=0.98\linewidth]{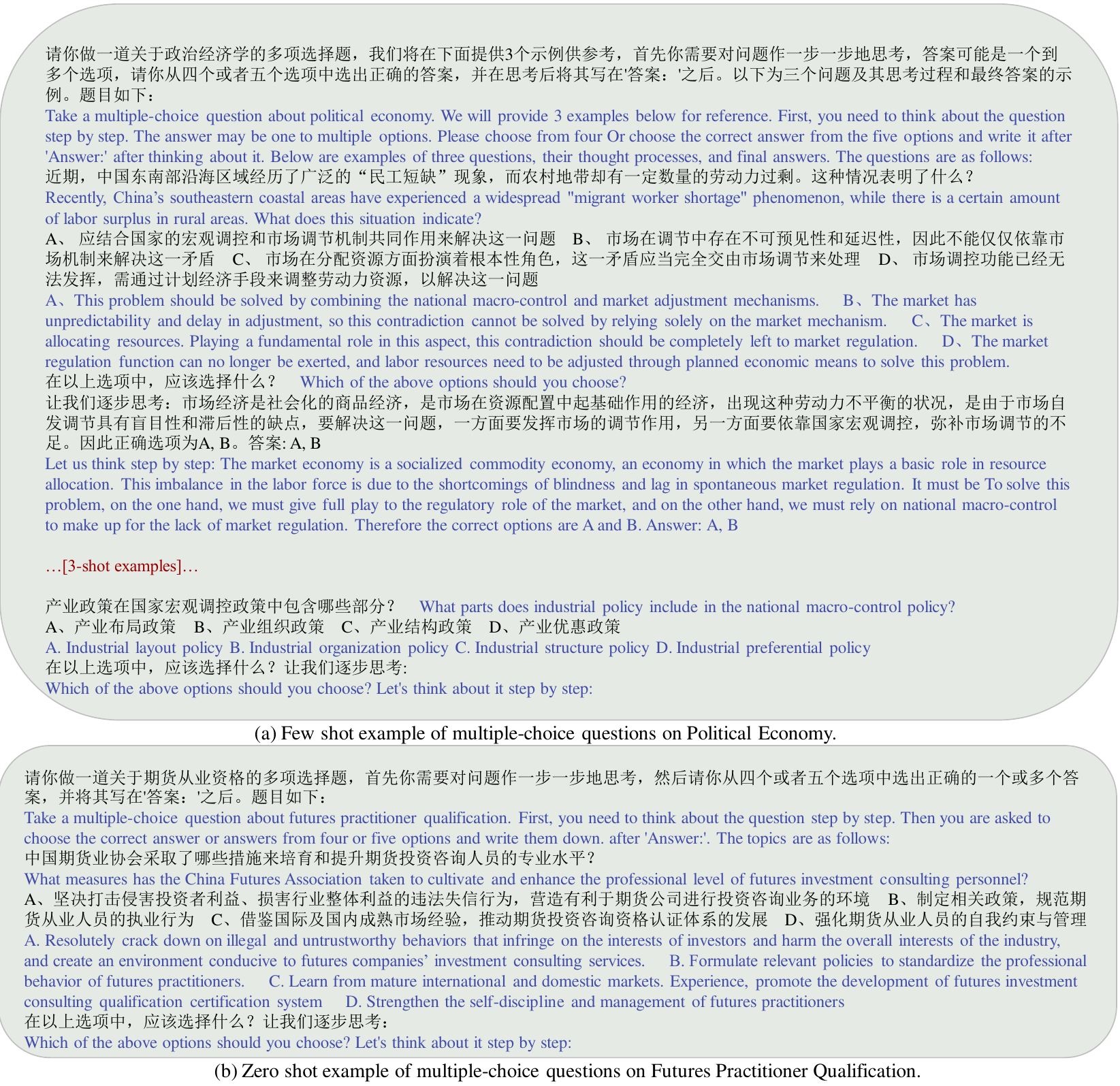}
	\caption{Examples of prompts in chain-of thought setting. English translations are shown in blue for better readability.}
	\label{appendix_fig3}
\end{figure}

We list the prompt examples utilized in the evaluation process, including zero-shot and few-shot in answer-only scenarios (~Figure\ref{appendix_fig1} and Figure\ref{appendix_fig2}), zero-shot and few-shot in chain-of-thought scenarios (~Figure\ref{appendix_fig3}).

\subsection{Data statistic}
In Table~\ref{appendix_tab4}, we enumerate the comprehensive statistical information of the dataset.

\begin{table}[t]
	\centering
	\caption{The detailed statistic of CFinBench dataset.}
	\small
	\setlength{\tabcolsep}{1.0mm}\begin{tabular}{lcccc}
		\toprule
		Category & Single-Choice  & Multiple-Choice & Judgment & All  \\
		\midrule
		\cellcolor[HTML]{EFEFEF}\textit{\textbf{Subject}} & \cellcolor[HTML]{EFEFEF}\textit{\textbf{3302}} & \cellcolor[HTML]{EFEFEF}\textit{\textbf{1889}} & \cellcolor[HTML]{EFEFEF}\textit{\textbf{3915}} & \cellcolor[HTML]{EFEFEF}\textit{\textbf{9106}}  \\  
		\midrule
		Political Economy  (\cc{政治经济学}) & 115 & 47 & 67 & 229  \\
		Western Economics (\cc{西方经济学}) & 268 & 46 & 212 & 526  \\
		Microeconomics (\cc{微观经济学}) & 295 & 29 & 221 & 545  \\
		Macroeconomics (\cc{宏观经济学}) & 21 & 117 & 294 & 432  \\
		Industrial Economics (\cc{产业经济学}) & 406 & 199 & 249 & 854  \\
		Public Finance (\cc{财政学}) & 167 & 86 & 156 & 409  \\
		International Trade (\cc{国际贸易学}) & 99 & 48 & 100 & 247  \\
		Statistics (\cc{统计学}) & 974 & 794 & 1846 & 3614  \\
		Auditing (\cc{审计学}) & 443 & 429 & 381 & 1253  \\
		Economic History (\cc{经济史}) & 248 & 63 & 133 & 444  \\
		Finance (\cc{金融学}) & 266 & 31 & 256 & 553  \\
		\midrule
		\cellcolor[HTML]{EFEFEF}\textit{\textbf{Qualification}} & \cellcolor[HTML]{EFEFEF}\textit{\textbf{14604}} & \cellcolor[HTML]{EFEFEF}\textit{\textbf{8879}} & \cellcolor[HTML]{EFEFEF}\textit{\textbf{5905}} & \cellcolor[HTML]{EFEFEF}\textit{\textbf{29388}}  \\ 
		\midrule 
		Tax Practitioner Qualification (\cc{税务从业资格}) & 1332 & 1544 & 464 & 3340  \\
		Futures Practitioner Qualification (\cc{期货从业资格}) & 2086 & 1396 & 1049 & 4531  \\
		Fund Practitioner Qualification (\cc{基金从业资格}) & 3892 & 118 & 536 & 4546  \\
		Real Estate Practitioner Qualification (\cc{地产从业资格}) & 503 & 511 & 660 & 1674  \\
		Insurance Practitioner Qualification  (\cc{保险从业资格}) & 1780 & 1220 & 903 & 3903  \\
		Securities Practitioner Qualification (\cc{证券从业资格}) & 2734 & 2041 & 1518 & 6293  \\
		Banking Practitioner Qualification  (\cc{银行从业资格}) & 258 & 173 & 266 & 697  \\
		Certified Public Accountant (CPA) (\cc{注册会计师}) & 2019 & 1876 & 509 & 4404  \\
		\midrule
		\cellcolor[HTML]{EFEFEF}\textit{\textbf{Practice}} & \cellcolor[HTML]{EFEFEF}\textit{\textbf{18824}} & \cellcolor[HTML]{EFEFEF}\textit{\textbf{13419}} & \cellcolor[HTML]{EFEFEF}\textit{\textbf{9802}} & \cellcolor[HTML]{EFEFEF}\textit{\textbf{42045}}  \\ 
		\midrule 
		Junior Auditor (\cc{初级审计师}) & 317 & 317 & 194 & 828  \\
		Intermediate Auditor (\cc{中级审计师}) & 237 & 223 & 197 & 657  \\
		Junior Statistician (\cc{初级统计师}) & 158 & 190 & 97 & 445  \\
		Intermediate Statistician (\cc{中级统计师}) & 259 & 400 & 195 & 854  \\
		Junior Economist (\cc{初级经济师}) & 2262 & 1496 & 655 & 4413  \\
		Intermediate Economist (\cc{中级经济师}) & 2547 & 1250 & 913 & 4710  \\
		Junior Banking Professional (\cc{初级银行从业人员}) & 2886 & 2075 & 1646 & 6681  \\
		Intermediate Banking Professional (\cc{中级银行从业人员}) & 2572 & 1550 & 1482 & 5604  \\
		Junior Accountant (\cc{初级会计师}) & 1654 & 1217 & 964 & 3835  \\
		Intermediate Accountant (\cc{中级会计师}) & 1252 & 858 & 700 & 2810  \\
		Tax Consultant (\cc{税务师}) & 934 & 1115 & 493 & 2542  \\
		Asset Appraiser (\cc{资产评估师}) & 1779 & 1690 & 896 & 4365  \\
		Securities Analyst (\cc{证券分析师}) & 1967 & 1038 & 1370 & 4375  \\
		\midrule
		\cellcolor[HTML]{EFEFEF}\textit{\textbf{Law}} & \cellcolor[HTML]{EFEFEF}\textit{\textbf{7695}} & \cellcolor[HTML]{EFEFEF}\textit{\textbf{5438}} & \cellcolor[HTML]{EFEFEF}\textit{\textbf{5428}} & \cellcolor[HTML]{EFEFEF}\textit{\textbf{18561}}  \\  
		\midrule
		Tax Law I (\cc{税法~I}) & 287 & 284 & 237 & 808  \\
		Tax Law II (\cc{税法~II}) & 283 & 323 & 238 & 844  \\
		Tax Inspection (\cc{税务稽查}) & 974 & 874 & 1664 & 3512  \\
		Commercial Law (\cc{商业法}) & 331 & 599 & 201 & 1131  \\
		Securities Law (\cc{证券法}) & 1009 & 106 & 693 & 1808  \\
		Insurance Law  (\cc{保险法}) & 69 & 57 & 42 & 168  \\
		Economic Law (\cc{经济法}) & 610 & 424 & 405 & 1439  \\
		Banking Law (\cc{银行业法}) & 2783 & 1360 & 1231 & 5374  \\
		Futures Law (\cc{期货法}) & 922 & 884 & 477 & 2283  \\
		Financial Law (\cc{金融法}) & 315 & 323 & 180 & 818  \\
		Civil Law (\cc{民法}) & 112 & 204 & 60 & 376  \\
		\midrule
		\cellcolor[HTML]{EFEFEF}\textit{\textbf{Total}} & \cellcolor[HTML]{EFEFEF}\textit{\textbf{44425}} & \cellcolor[HTML]{EFEFEF}\textit{\textbf{29625}} & \cellcolor[HTML]{EFEFEF}\textit{\textbf{25050}} & \cellcolor[HTML]{EFEFEF}\textit{\textbf{99100}}  \\  
		\bottomrule
	\end{tabular}
	\label{appendix_tab4}
\end{table}

\subsection{Limitations and potential negative societal impacts}
For limitations, ours focuses on Chinese financial system and is therefore not suitable for assessing financial knowledge in other countries. The proposed CFinBench provides an important basis for evaluating the LLMs' mastery of Chinese financial knowledge. However, since we have open sourced the questions and answers of the validation split, if they are improperly used to train the large language model, the accuracy of the model may be falsely high. Therefore, the results on CFinBench are only a reference. The true quality of the model depends on the performance of the user in the practical scenario.

\subsection{License}
The proposed benchmark is under the terms of the Apache-2.0 license\footnote{https://www.apache.org/licenses/LICENSE-2.0}. We encourage code sharing and respect the copyright of the original author, and also allow code modification and redistribution (as open source or commercial software).

\end{document}